# Generative Adversarial Networks for Labelled Vibration Data Generation

Furkan Luleci[1]; F. Necati Catbas[1*], Ph.D., P.E.; and Onur Avci[2], Ph.D., P.E.

[1]Department of Civil, Environmental, and Construction Engineering, University of Central Florida,

Orlando, FL, 32816, USA

[2] Department of Civil, Construction, and Environmental Engineering, Iowa State University,

Ames, IA, 50011, USA

**ABSTRACT**

As Structural Health Monitoring (SHM) being implemented more over the years, the use of operational modal analysis of civil structures has become more significant for the assessment and evaluation of engineering structures. Machine Learning (ML) and Deep Learning (DL) algorithms have been in use for structural damage diagnostics of civil structures in the last couple of decades. While collecting vibration data from civil structures is a challenging and expensive task for both undamaged and damaged cases, in this paper, the authors are introducing Generative Adversarial Networks (GAN) that is built on the Deep Convolutional Neural Network (DCNN) and using Wasserstein Distance for generating artificial labelled data to be used for structural damage diagnostic purposes. The authors named the developed model "1D W-DCGAN" and successfully generated vibration data which is very similar to the input. The methodology presented in this paper will pave the way for vibration data generation for numerous future applications in the SHM domain.

*E-mail: catbas@ucf.edu

1) **INTRODUCTION**

Tracking the current and projected increments in the catastrophic events due to climate change, aging, and deteriorating civil structures due to the human and environmental factors has made Structural Health Monitoring (SHM) a useful technology for decision making. Therefore, in the context of SHM, structural damage diagnostics and prognostics on the collected sensor data from the civil structures have become more and more important for the condition assessment and evaluation of the civil structures. Thus, early identification and prognosis on the faulty data can save costly structural failures and more importantly, human lives. To that end, there have been several parametric and non-parametric damage diagnostic and prognostic studies introduced in the literature. Also, with the emergence of Machine Learning (ML) and Deep Learning (DL) methods, using them in the SHM field has gained a lot of attention due to yielding successful and robust results on the trained input data [1-23].

1.1) Problems

To be able to work with ML and DL algorithms, training the model on the inputted data is essential for the supervised and semi-supervised methods. At this point, among other parameters in the model, the amount of data to be trained is one of the crucial factors of the model's success. Thus, it is important to have as much data as possible. However, in the SHM field, collecting data through sensors from the structure has several challenges, such as getting permission from authorities to work on the structure, traffic closures, setting up an expensive SHM system, and needing skilled experts on the field. Additionally, finding useful data (data that contains damage features) from civil structures is a rare opportunity. This is where the ML and particularly DL models can be highly beneficial as they provide successful results with large amount of data. For this reason, there are studies that attempted to generate data that is similar to the trained input.

1.2) Background

In 2014, Ian Goodfellow and his colleagues introduced the Generative Adversarial Networks (GAN) which are based on a game theory where two neural networks play a zero-sum game, thus the two separate networks compete to be better. As a result, the converged model can produce very similar results to the inputted data [24]. Subsequently, there have been other studies aimed to improve the GAN's training problems such as Deep Convolutional Generative Adversarial Networks (DCGAN) [25], Wasserstein GAN (W-GAN) [26]. GANs are heavily used in the computer vision field particularly producing very real looking human face images. Also, there are some studies on 1-D data generation and reconstruction using GANs [27-41]. In the SHM for civil structures field, few studies exist on 1-D data generation and reconstruction using GANs [42-45], but none of them focus on structural damage detection.

1.3) Objective of the study

The objective of this paper is to produce a useful vibration dataset by using an improved GAN model, which is a GAN variant built on a Deep Convolutional Neural Network (DCNN) architecture that uses the Wasserstein distance metric. Thus, we name our model 1D-WDCGAN since it is used on a 1-D vibration data. Thereby, the proposed study investigates that during building an ML or DL based damage diagnostic model where the amount of useful vibration dataset is not sufficient to train the model, the successfully generated artificial dataset from 1-D W-DCGAN can be used to train the model.

2) **WORKFLOW**

The general workflow can be summarized in this order: (1) Data preprocessing, (2) 1-D W-DCGAN model building, (3) Training and fine-tuning, and (4) Evaluation and interpretation of the results. Also, other types of GAN models are tried (GAN, DCGAN, W-DCGAN with Gradient Penalty loss), but W-DCGAN gave the most robust and accurate results along with more stable training.

2.1) Data and Equipment

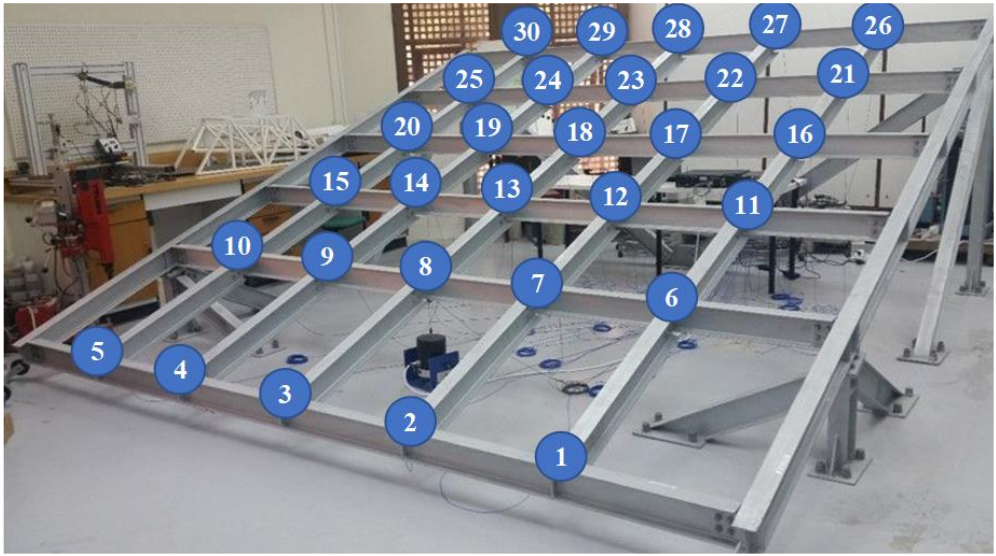

Figure 1. Steel frame grandstand simulator [9]

In study [9], the authors conducted a successful 1-D CNN-based level 1-2-3 (detection, localization, and quantification) damage diagnostic of a steel frame of a grandstand simulator (Figure 1). They collected vibration datasets at each joint for 256 seconds at a 1024 Hz sampling rate (total of 262,144 samples) from a created 30 different damaged and 1 undamaged scenarios where each damage scenario consists of bolt loosening at a joint respectively. This study aimed to create a damaged tensor, therefore only one column of vibration dataset from damage scenario (1) at joint (1) is used, which is a 256 second vibration dataset when joint (1) is induced with bolt loosening. The dataset provided in this study is open access which can be found in this reference [46]. Furthermore, the 1-D W-DCGAN is trained on a desktop PC that is equipped with 16 GB RAM DDR4 2933Mhz and NVIDIA GeForce RTX 3070 8GB GDDR6 graphic card.

2.2) Data Preprocessing

Before training the CNN models, normalizing the inputs is common practice to make the train and test datasets in the same scale for the model to learn and predict better. Also, during the dot product of weights in the front and backpropagation through the network, it helps the model to be more accurate and take less computation time. However, only a single operation is necessary for the GANs, which is only training, thus normalization is not needed unless the input signal has spikes, otherwise different scaled weight propagations can lower the model quality during the training. Additionally, the used 1-D W-DCGAN consists of batch and instance normalization layers which help the batches of data to be normalized during the training. After several trials with normalized input and raw input, the result did not noticeably change and in fact, it is believed that the 1-D W-DCGAN model can learn the spatial and temporal features better if the raw signal is fed in the model. Thus, the input is not normalized, and the raw 262,144 length of damage tensor is used as an input.

2.3) Model Architecture

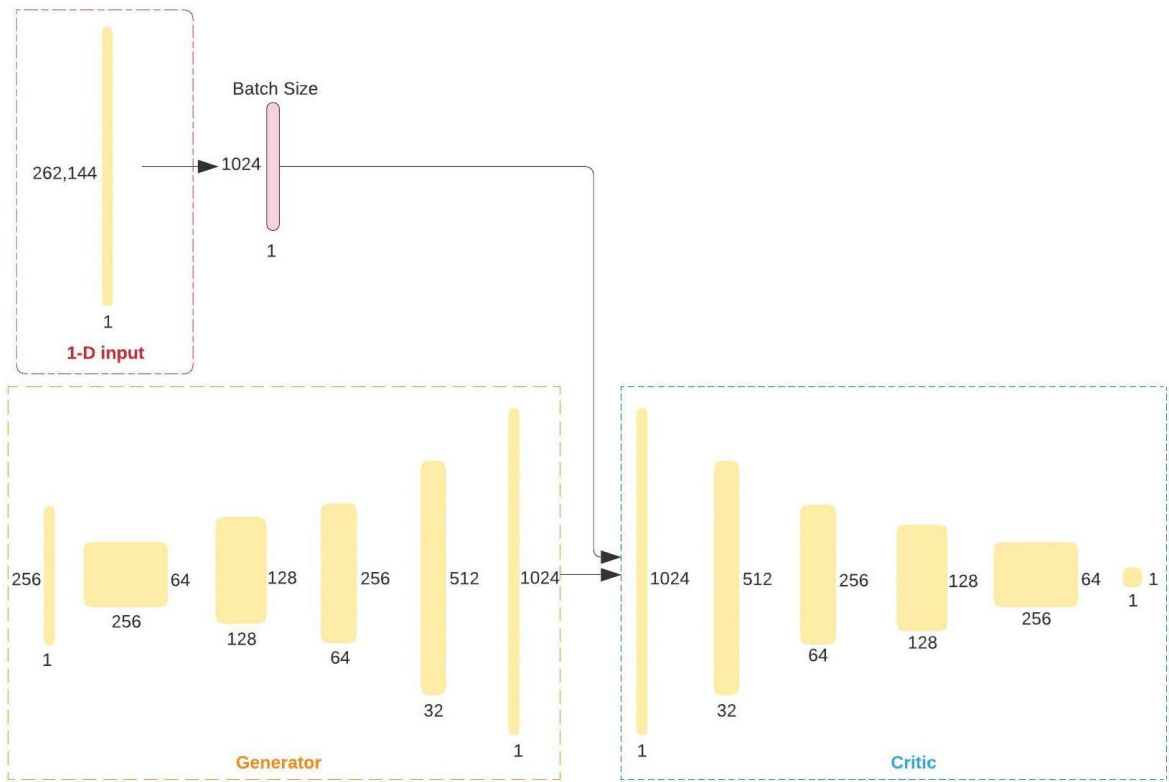

Figure 2. 1-D W-DCGAN model architecture

In the model, the generator takes the [1 x 256] dimensional noise tensor (z) and pass it through 5 1-D transpose convolutions, then followingly the [1 x 1024] vibration tensor is created. This is the point where both the created tensor from generator and the sampled batches of [1 x 1024] from the 1-D input data (x) enters the discriminator which is named as critic in the W-GAN paper. Then the critic takes the inputted 1-D tensor and after 5 1-D convolutions, it yields the decision score of the Critic to backpropagate and optimize the network afterwards. The readers are directed to the GAN [24], DCGAN [25], and W-GAN [26] papers for more details about the original models and the training process. After several trials, the best architecture is formed as shown in Figure 2. Additionally, the generator starts the 1-D transpose convolutions with filter = 64, stride = 2, padding = 0. Then, continues for the rest with filter = 4, stride = 2, padding = 1. The critic takes the same filter, stride, and padding values in reverse but the last 1-D convolution layer takes filter = 64, stride = 2, padding = 0. Note that due to the limited specifications of the used PC, the maximum number of length of features in the generated dataset could be 1024. That means, after the training, the model is going to learn the 262,144-length of dataset and generate a 1024-length dataset of variation of the input.

2.4) Fine-tuning for Training

Training GANs and its variants are notoriously the most challenging ones among other DL models. The most common challenges are: powerful discriminator over the generator; high oscillation and unstable training thus no convergence; mode collapse; in other words generator always creates the same output; diminishing gradients – losing gradients through the network with more epoch. A couple of approaches have been used to tackle these challenges and observed positive effects: Using W-GAN significantly helps in more stable and less sensitive training progress and leads to convergence with a trade-off with more time; Using dropout in the critic helped to reduce the capacity of the critic and overfitting the network; a decaying random Gaussian noise is added to the real training 1-D input to slow down the critic's learning progress. Furthermore, after many trials of different values of training, the learning rate $1\text{x}10^{-5}$ yielded the best results. The number of epoch is decided based on the loss values of the generator, the critic, and the value of Fréchet Inception Distance (FID) score as they approach zero; and after many trials, the epoch number 45 is found to yield the best results. As can be seen in Figure 3, the loss functions and similarity index, FID score, are converging to zero axis. Especially the FID score reduced from 0.0503 to 0.0013. For a baseline reference, in one study on MNIST data, the authors used the FID score to evaluate their model and the FID score was reduced from around 250 to 50 levels [49]. Considering that it can be concluded that our model can generate very realistic-looking datasets. The FID score is explained in detail in the next section. Also, using the AdamW optimizer in both the generator and the critic, 1-D batch normalization in the generator, and 1-D instance normalization in the critic yielded the best results in training.

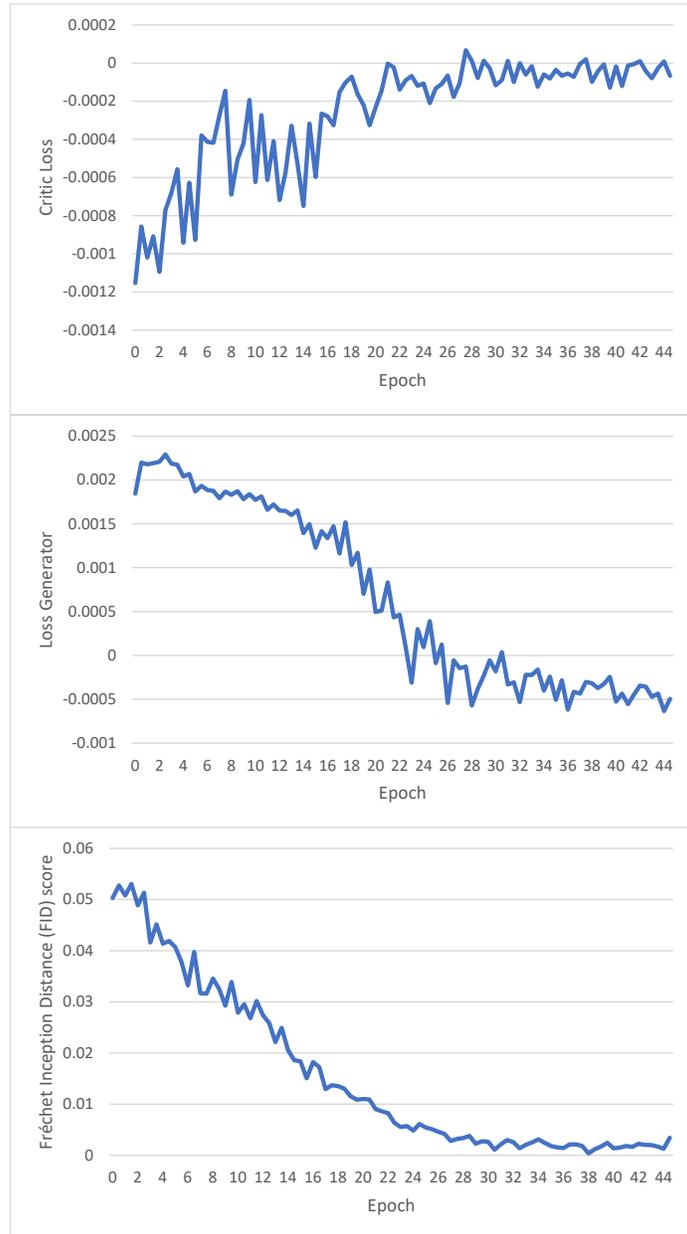

Figure 3. From top to bottom respectively Critic Loss, Generator Loss, FID Score values

2.5) Evaluation and Interpretation of the Results

Like other DL models, GANs are initially used on image-based applications when they first came out. That is why among other evaluation metrics, the most reliable one for the evaluation of results is visually comparing the generated fake image with the real one. However, this might not be an easy method for 1-D data as doing it for 2-D images also visual evaluation has limitations such as introducing subjectivity for a different viewer and a limited number of data can be reviewed in limited time. Additionally, GANs do not have an objective function or measure, which makes it challenging for evaluation of the model, unlike other DL models. As there are several methods to compare performance, there is no consensus in the DL field yet as to which measure works best. The reader is directed to the reference [47] for a comparison of evaluation measures of GANs. One of the most effective and most used quantitative evaluation methods was proposed in 2018 for evaluating the performance of GANs [48]. The FID is introduced as an improvement over the Inception Score which lacks capturing the training input (real) to the generated output. The FID score calculates the mean and standard deviation of the real and generated data and the lower the score means, the more similar distributions are. The FID formula used in this study is provided in Equation 1.

$$FID = (|\mu_x - \mu_y|)^2 + (\sigma_x - \sigma_y)^2 \tag{1}$$

Where $\mu_x$ and $\mu_y$ refer to the mean and $\sigma_x$ and $\sigma_y$ refer to the standard deviation of the real and the generated data, respectively. Note that the FID calculations are executed between two [1 x 1024] vibration column tensors $[a11_b]$ and $[a11_f]$, $[a11_b]$ is where the column tensor of 1024-length randomly batch-sampled from the real input column tensor of 262,144-length and $[a11_f]$ is where the produced fake column tensor of 1024-length from the generator.

After the training, to calculate the FID score between the real input and the generated ones, 256 number of 1024-column tensors are generated and 256 number of 1024-column tensors are batch-sampled from the real input. The output of this is a 65,536-column tensor of FID scores where after the FID calculation of 256 real column tensors and 256 amount of fake column tensors produced a 65,536-column tensor. The lowest FID score is found 0.00002 and the highest one is found 0.00387. Considering that in the training, the FID score is started with 0.0503, it can be concluded that the model performed the training process on the input data exceptionally well and can generate very similar data. Also, the probability density distribution of FID scores of the 65,536-column tensor is plotted in Figure 4 and the values are accumulated around 0.001 which is a reduction from 0.0503 by 50 times and can be concluded that statistically the generated samples are very similar to the real ones.

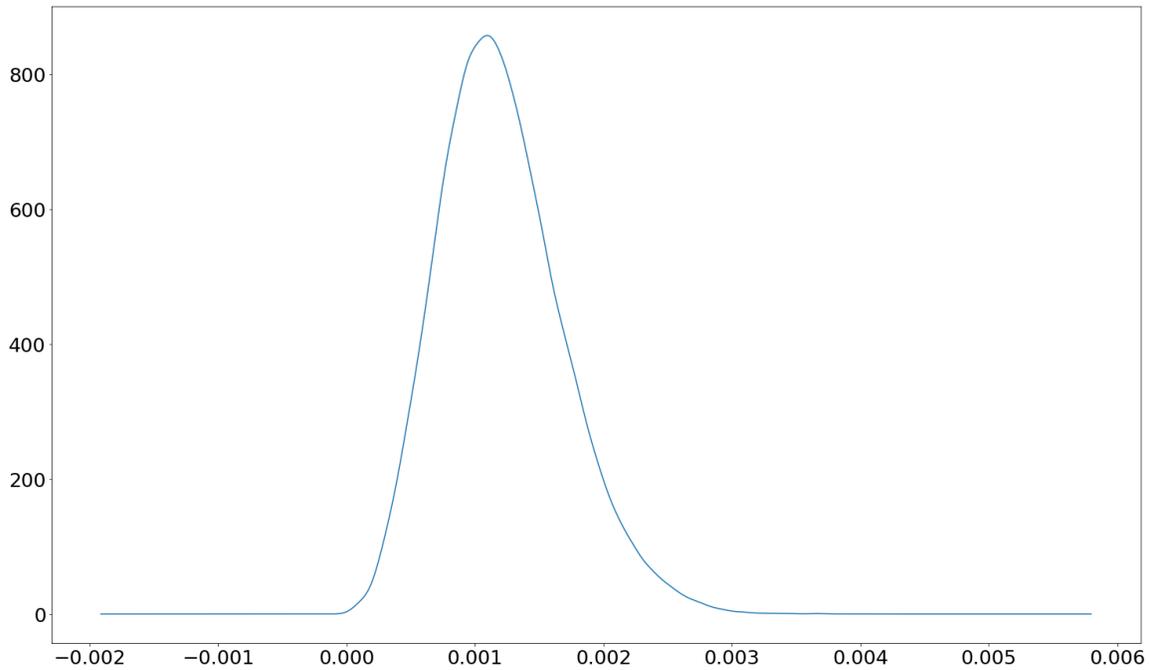

Figure 4. Probability density distribution of the FID scores

For manual evaluation of the results, among 256 generated fake and 256 batch-sampled column tensors, a one with high, medium, and low FID score (the FID scores are displayed on the figure corresponding to each signal pair) vibration column tensors are plotted in Figure 5 and there is some consistency between the tensor pairs. However, as mentioned before, visual evaluation for the performance of the model for 1-D data is not as effective as implementing it on images. That's why quantitative evaluation methods like the FID can be more reliable as it is based on the statistical measure. Furthermore, the same plotted tensors in Figure 5 are box plotted and shown in Figure 6. There is a very slight divergence (less than 0.01) in mean lines, median lines, quartile boundaries, and whisker values. However, as the goal of the GANs is not to produce the same exact input but a slight variance of it and provide sufficient enough similar output to the real input to fool the critic or discriminator, the results can be concluded as very good.

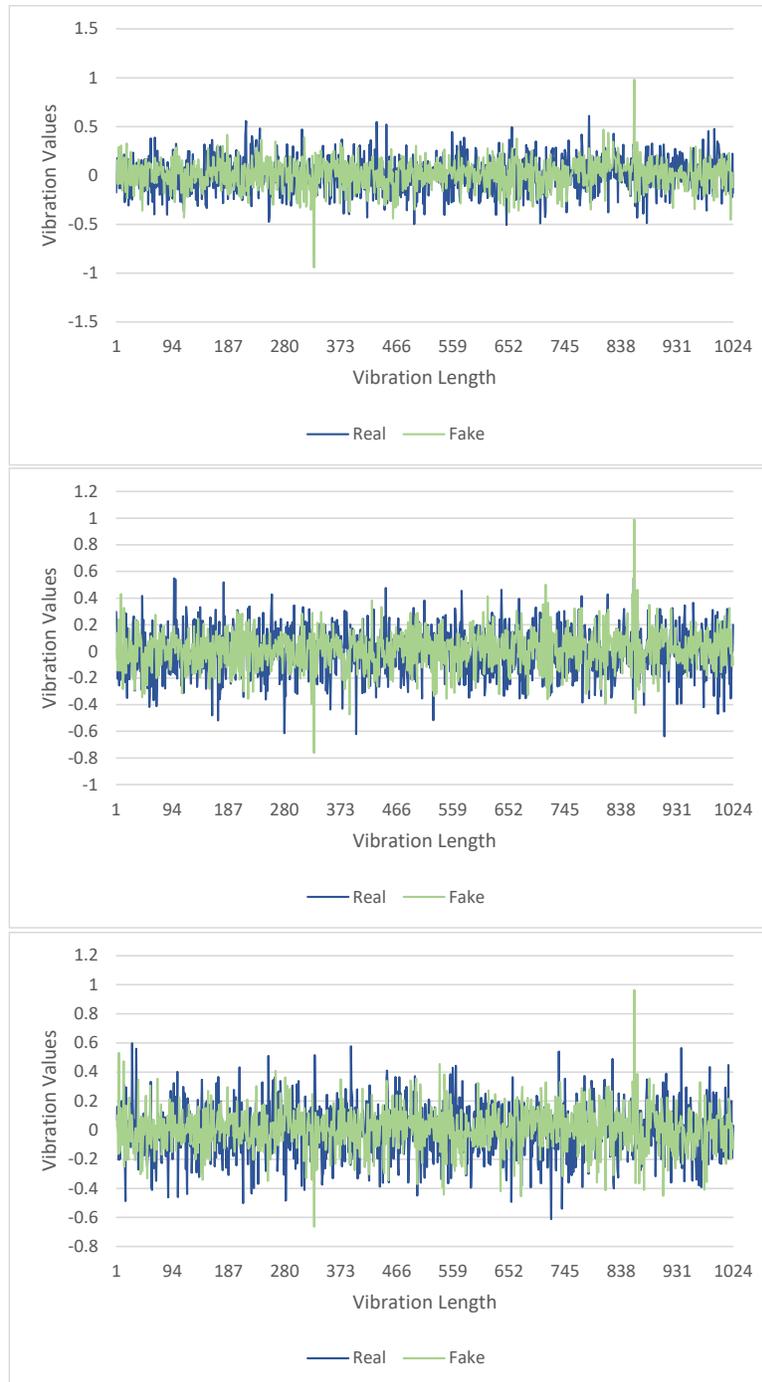

Figure 5. Generated 1024-length tensors: Top to bottom FID Scores are 0.00002, 0.00118, and 0.00387

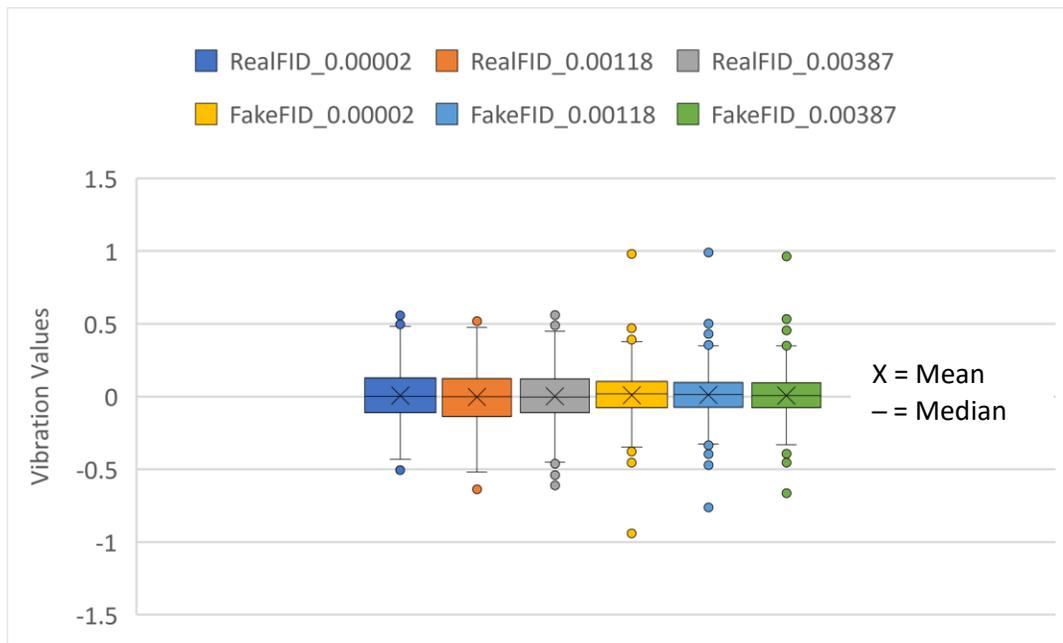

Figure 6. Box plot of the plotted vibration tensors in Figure 5

### 3) CONCLUSION

In this study, by building a 1-D W-DCGAN, vibration column tensors are generated whose input is a 256 seconds (262,144 samples) vibration tensor under given a damage scenario at joint 1. Due to the limited PC specs, the model is able to produce 1 second (1024 samples) of vibration column tensor by batch-sampling the input. The quantitative evaluation of the results by using the FID score yielded outstanding results as it got very close to zero. Although there is a very slight divergence in the generated data to the real input in the box plot visualization of picked high, medium, and low FID score tensors, it is interpreted that the small difference is an indication of slight variance of the real input as the intention is not creating the same exact input. The qualitative evaluation shows consistency between the plotted signal pairs; however, it is not trivial as evaluating the 2-D images on which the GANs are implemented first.

The goal of this study was to generate useful signals (in the SHM field this is mostly damage associated data) in the case of having an insufficient amount of data (insufficient damage data is a very common challenge in the SHM field). Thus, by enhancing the dataset with 1-D W-DCGAN and later using it to train an ML or DL model for vibration-based damage diagnostics, the prediction results of the ML or DL model could be more accurate. Yet, to validate the results of this study, it is essential to try the generated data from a GAN model on an ML or DL model.